\documentclass{article}

\usepackage[final,nonatbib]{neurips_2024_ml4ps}

\usepackage[utf8]{inputenc} 
\usepackage[T1]{fontenc}    
\usepackage{hyperref}       
\usepackage{url}            
\usepackage{booktabs}       
\usepackage{amsfonts}       
\usepackage{nicefrac}       
\usepackage{microtype}      
\usepackage{xcolor}         

\usepackage{comment}
\usepackage{graphicx} 
\usepackage{tabularx}
\usepackage{amsmath}
\usepackage{bbm} 
\usepackage[
backend=biber,
style=numeric,
sorting=none
]{biblatex}
\title{A bibLaTeX example}
\usepackage[style=numeric,sorting=none]{biblatex}
\title{3D Cloud reconstruction through geospatially-aware Masked Autoencoders}

\author{%
  Stella Girtsou\\
  National Observatory of Athens \\
  National Technical University of Athens \\
  \texttt{girtsou.s@gmail.com} \\
     \And
 Emiliano Diaz Salas-Porras\\
 Universitat de València \\
    \texttt{emdiazsal@gmail.com} \\
  \AND
  Lilli Freischem \\
  University of Oxford \\
  \texttt{lilli.freischem@physics.ox.ac.uk} \\
  \And
  Joppe Massant \\
  Royal Belgian Institute of Natural Sciences \\
  \texttt{joppe.massant@gmail.com} \\
  \And
  Kyriaki-Margarita Bintsi \\
  Imperial College London \\
  \texttt{m.bintsi19@imperial.ac.uk} \\
  \And
  Guiseppe Castiglione \\
  University of Sussex \\
  \texttt{g.m.a.castiglione@gmail.com} \\
  \And
  William Jones \\
  University of Oxford \\
  \texttt{william.jones@physics.ox.ac.uk} \\
  \And
  Michael Eisinger \\
  European Space Agency \\
  \texttt{Michael.Eisinger@esa.int} \\
  \And
  Emmanuel Johnson \\
  CSIC-UCM-IGEO \\
  \texttt{jemanjohnson34@gmail.com} \\
  \And
  Anna Jungbluth \\
  European Space Agency \\
  \texttt{anna.jungbluth@esa.int} \\
}

\begin{document}

\maketitle

\begin{abstract}
    Clouds play a key role in Earth's radiation balance with complex effects that introduce large uncertainties into climate models. Real-time 3D cloud data is essential for improving climate predictions. This study leverages geostationary imagery from MSG/SEVIRI and radar reflectivity measurements of cloud profiles from CloudSat/CPR to reconstruct 3D cloud structures. We first apply self-supervised learning (SSL) methods—Masked Autoencoders (MAE) and geospatially-aware SatMAE-on unlabelled MSG images, and then fine-tune our models on matched image-profile pairs. Our approach outperforms state-of-the-art methods like U-Nets, and our geospatial encoding further improves prediction results, demonstrating the potential of SSL for cloud reconstruction.
\end{abstract}

\section{Introduction}
Clouds play a crucial role in Earth's radiation balance by reflecting sunlight (cooling) and absorbing heat (warming) \cite{wang1995determination}. 
As climate change progresses their complex interactions increase uncertainties in climate models. Enhanced global, real-time 3D cloud data can help reduce these uncertainties, improving climate predictions and decision-making.
Satellite remote sensing has transformed climate research by providing global data. 
The European Space Agency's recently launched EarthCARE mission aims to improve our understanding of cloud dynamics through imaging, radar, and lidar instruments \cite{wehr2023earthcare}. 
Over the last decades NASA's CloudSat mission has provided valuable measurements of cloud vertical profiles using radar, but is limited by infrequent revisits (every 16 days) and a narrow swath (1.4 km). Additionally, with its sun-synchronous orbit, CloudSat passes over the equator at the same local time, limiting observations to "snapshots" of the atmosphere at specific times of day.
In contrast, imaging instruments take measurements across wider fields-of-view and with higher temporal resolution, but they only offer a 'top-down' perspective and do not directly measure atmospheric profiles.
However, combining images in different spectral channels with overlapping measurements of atmospheric profiles allows extrapolation of vertical profiles beyond the radar track. 
Barker et al. \cite{Barker2011, Qu2023} developed an algorithm to extend EarthCARE profiles to 3D through intensity pixel-matching. 
Recent work \cite{Bruning2024-db, Jeggle-2023, Leinonen2019} has used ML-based methods (e.g. U-Nets, CGANs, Linear Regression, Random Forests, XGBoost) to estimate vertical cloud information from 'top-down' measurements. 
Notably Br\"uning et al. \cite{Bruning2024-db} trained a Res-UNet to fuse satellite images from the Meteosat Second Generation (MSG) Spinning Enhanced Visible and InfraRed Imager (SEVIRI) with CloudSat cloud profiling radar (CPR) reflectivity profiles, reconstructing 3D cloud structures. 
For all approaches, model training required precise spatial and temporal alignment between the data sources. 
Due to the limited overpasses of the radar satellites (Figure \ref{Figure-Pipeline}b), profile measurements are much sparser than available imagery (for comparison, MSG/SEVIRI generates 40 TB of image data per year, while CPR produces 150 GB annually). 
Recent advances in self-supervised learning (SSL) have shown promise in pre-training models on large, unlabeled datasets, yet their application to cloud studies remains under-explored. 
In this work, we apply SSL methods—Masked Autoencoders (MAE, \cite{he2021maskedautoencodersscalablevision}) and geospatially-aware Masked Autoencoders (based on SatMAE, \cite{cong2023satmaepretrainingtransformerstemporal})—to multispectral MSG/SEVIRI data from 2010. 
We then fine-tune the pre-trained models for the task of 3D cloud reconstruction using matched image-profile pairs. 
Our results show that pre-training consistently improves performance for this task, especially in complex regions like the tropical convection belt. 
Pre-trained models with geospatial awareness (i.e., time and coordinate encoding) especially outperform randomly initialized networks and simpler U-Net architectures, leading to improved reconstruction results. 
The code will be made available upon acceptance.

\section{Data \& Method}

\textbf{Input Imagery.}
We use radiance data from MSG/SEVIRI in 11 spectral channels as model input. 
MSG takes measurements from a geostationary orbit, centered at 0° longitude and latitude, and covers a ±80° field-of-view, which we limit to ±45° to avoid data quality issues.
New images are captured every 15 mins, offering continuous monitoring at a spatial resolution of 3\,km at the sub-satellite point.
From the 40 TB of annually available data, our pre-training dataset comprises hourly cadence data split monthly into days 1-10 (training), days 21-22 (validation), and days 25-27 (testing). 
We split each full-disk image into non-overlapping patches of 256 x 256 pixels, resulting in 1.2 million (8.6 TB) patches.
To increase training speed and reduce computational load, we randomly sample a different 10\% of our patches each epoch during pre-training. 
With this sub-sampling, convergence during pre-training remains smooth and stable.

\textbf{Target Profiles.} We use radar reflectivity profiles measured via the CPR onboard the CloudSat satellite as our target. 
We spatially and temporally align CloudSat/CPR radar reflectivity from 2010 with MSG/SEVIRI images, and restrict our dataset to Cloudsat tracks with at least 20 \% cloud cover.
This results in approximately 47 thousand image-profile pairs consisting of a 11 x 256 x 256 pixels image and a geo-referenced CloudSat profile with 125 vertical bins, spanning roughly 30 km into the atmosphere. 
We crop 25 (10) height levels at the bottom (top) of each profile to remove measurement artifacts and reduce clear sky data \cite{Leinonen2019, Bruning2024-db}. 
The length of each profile varies due to the intersection between CloudSat's trajectory and the SEVIRI image.
Similar to pre-training, we use days 1-19, 23-24, and 28-31 for training, days 20-22 for validation, and days 25-27 for testing. 
Since clouds evolve over minutes to hours, a temporal split based on multi-day groupings ensures minimal data leakage. 
We randomly sample 50\% of all image-profile pairs per epoch.

\begin{figure}
    \centering
    \includegraphics[width=0.95\linewidth]{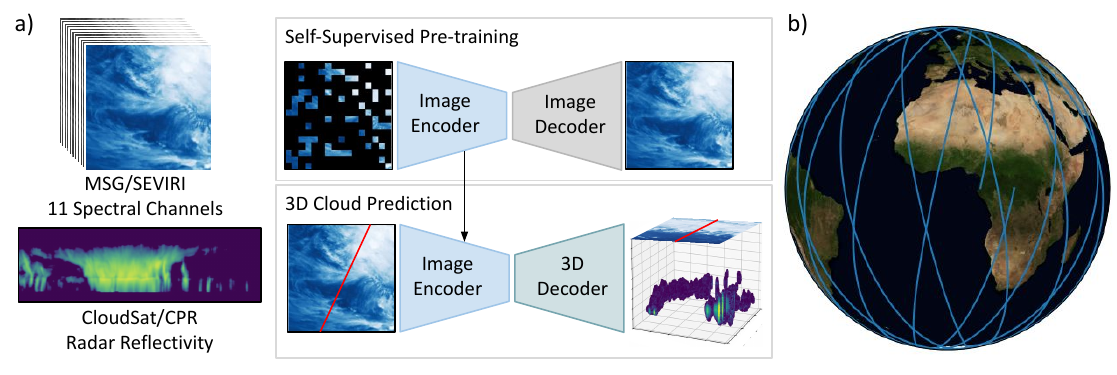}
    \caption{(a) Proposed pipeline: We use MAE to pre-train encoders on unlabelled MSG/SEVIRI images by reconstructing missing information, then fine-tune these encoders with a smaller dataset of image-profile pairs to derive 3D cloud structures. (b) The ground track of 1 day of CloudSat orbits.}
    \label{Figure-Pipeline}
\end{figure}

\textbf{Models \& Training.} For pre-training, we use MAEs with Vision Transformer (ViT) backbones to tokenize MSG/SEVIRI images and learn local features and their spatial relationships through positional encodings. We chose MAEs for their potential to handle missing data, and ViTs for their self-attention mechanisms and ability to relate local and global image features - all important for Earth Observation satellite data.
We tested two ViT variants: a \emph{base} ViT-B (90M parameters) and a \emph{small} ViT-S (26M), encoding $11 \times 256 \times 256$ images with $768$ and $384$ hidden channels, respectively. 
Image tokenization was performed at 8x8 and 16x16 pixel resolutions, with 75\% of tokens masked for reconstruction using multi-layer decoders \cite{he2021maskedautoencodersscalablevision}.
Given our geospatial data, we also adapted the SatMAE framework \cite{cong2023satmaepretrainingtransformerstemporal}, adding temporal encodings (date and time) and spatial coordinates (latitude, longitude) to enhance geospatial awareness. 
Although we tested single-timestamp inputs, our model can also handle time series of images.
After self-supervised pre-training, we replaced the decoder with a regression head using transposed convolutional layers, and fine-tuned the models for our 3D cloud reconstruction task. 
Importantly, the final output of the decoder reconstructs 90 x 256 x 256 (H x W x L) volumes. 
Although the model generates full 3D cloud maps from MSG/SEVIRI images, the loss is only computed on the one vertical profile slice of the target CloudSat overpass, following the approaches in \cite{Jeggle-2023, Bruning2024-db}. 
Since this is a poorly constrained problem, we ensure that (1) the model does not have access to the overpass mask, i.e. it does not know where it will be tested, and (2) the target CloudSat overpass can be of variable length and through any part of the input image. 
This forces the model to reasonably reconstruct 3D volumes anywhere in the output cube. 
An overview of our pipeline is shown in Figure \ref{Figure-Pipeline}.
We compare our SSL approach to the U-Net presented by Br\"uning et al. \cite{Bruning2024-db}, which also reconstructs a 3D volume.

\section{Results}
\label{results}
\begin{figure}[t!]
    \centering
    \includegraphics[width=0.85\linewidth]{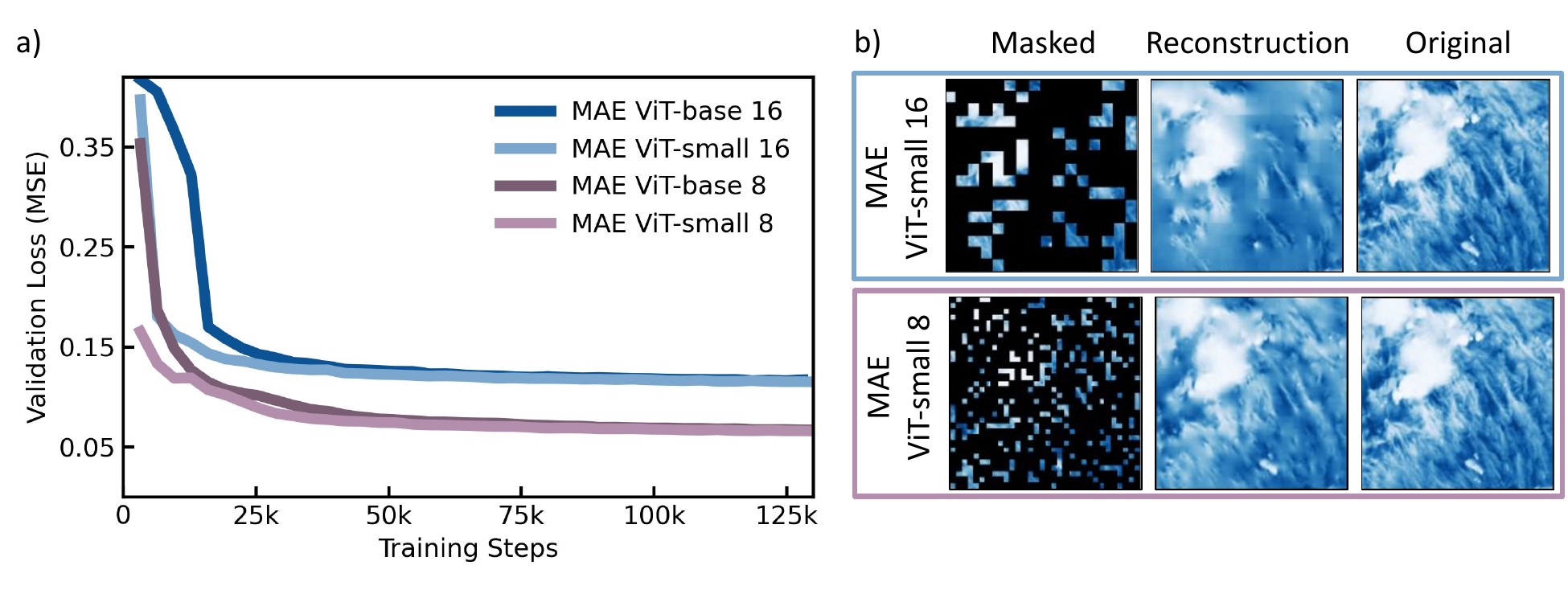}
    \caption{(a) Validation Loss for the different size MAEs. (b) Visualization of the masked, reconstructed and original images for MAEs based on 8- or 16-pixel tokenization.}
    \label{Figure-MAE-Reconstruction}
\end{figure}

\textbf{Pre-training.}
We experimented with different sizes of ViT backbones, ViT-base (90M) and ViT-small (26M), and tokenization schemes. 
As shown in Figure \ref{Figure-MAE-Reconstruction}, smaller ViT tokens yield lower losses and finer-grained detail in the image reconstruction.
Since the small and base MAE models exhibit comparable performance, we chose the smaller ViT for its reduced computational demands.

\begin{table}[t]
  \caption{Comparison of metrics. The values show the mean and standard deviation across the test set.}
  \label{Table-Metrics-Comparison}
  \centering
  \small
  \begin{tabular}{lrrr}
    \toprule
     & U-Net & MAE ViT-small 8  & SatMAE ViT-small 8 \\
    \midrule
    \textbf{reflectivity} \\
    \midrule
    RMSE (dBZ) $\downarrow$ & 3.73 $\pm$ 1.97  & 3.22 $\pm$ 1.68 & 3.18 $\pm$ 1.67                    \\ 
    \% Error $\downarrow$ & 1.28 $\pm$ 2.51  & 0.72 $\pm$ 1.00 & 0.72 $\pm$ 1.01                    \\ 
    Peak signal-noise ratio (PSNR) $\uparrow$ & 24.39 $\pm$ 4.15 & 25.78 $\pm$ 4.47 & 25.92 $\pm$ 4.48                   \\ 
    Structural similarity index measure (SSIM) $\uparrow$ & 0.82 $\pm$ 0.09  & 0.84 $\pm$ 0.09 & 0.84 $\pm$ 0.09                    \\
    \midrule
    \textbf{segmentation} \\
    \midrule
    f1-score $\uparrow$ & 0.845 $\pm$ 0.083  & 0.874 $\pm$ 0.092 & 0.880 $\pm$ 0.087                    \\
    f1-score weighted $\uparrow$ & 0.290  $\pm$ 0.098  & 0.382 $\pm$ 0.135 & 0.381  $\pm$ 0.142
    \\
    \bottomrule
  \end{tabular}
\end{table}

\textbf{Fine-tuning.} To evaluate the benefit of self-supervised pre-training, we compare models that were trained from scratch, or fine-tuned with frozen or unfrozen backbones. 
As shown in Figure \ref{Figure-MAE-Comparison} in the Appendix, pre-training brings substantial benefit, especially when the backbone is unfrozen. 
The unfrozen MAE outperformed all models in validation loss, achieved better PSNR and demonstrated faster, smoother convergence than the U-Net. 

\begin{figure}
    \centering
    \includegraphics[width=0.9\linewidth]{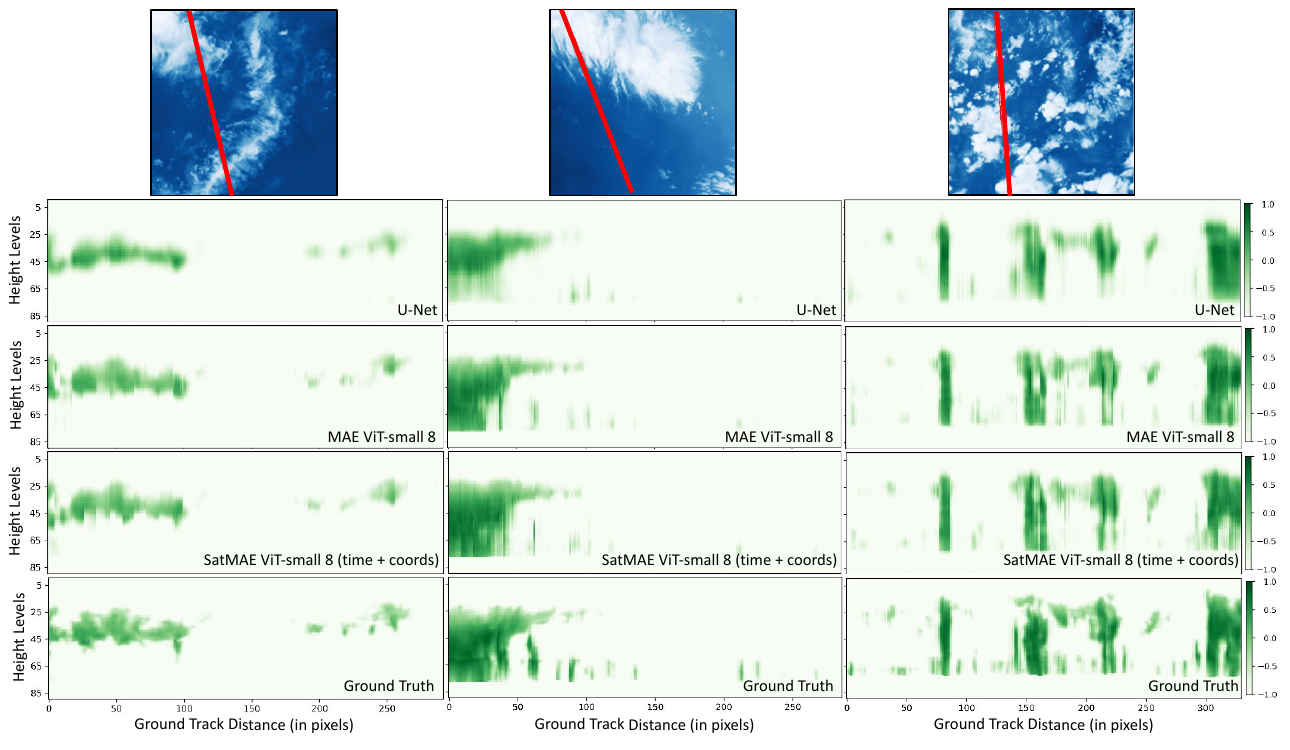}
    \caption{Comparison of example MSG/SEVIRI input channels ($7.35~\mu\text{m}$) with CloudSat overpasses (red lines), and corresponding (normalized) radar reflectivity profiles. All models were trained for 50 epochs, which resulted in the highest perceptual quality.}
    \label{Figure-Predictions}
\end{figure}

\begin{figure}[htb]
    \centering
    \includegraphics[width=0.95\linewidth]{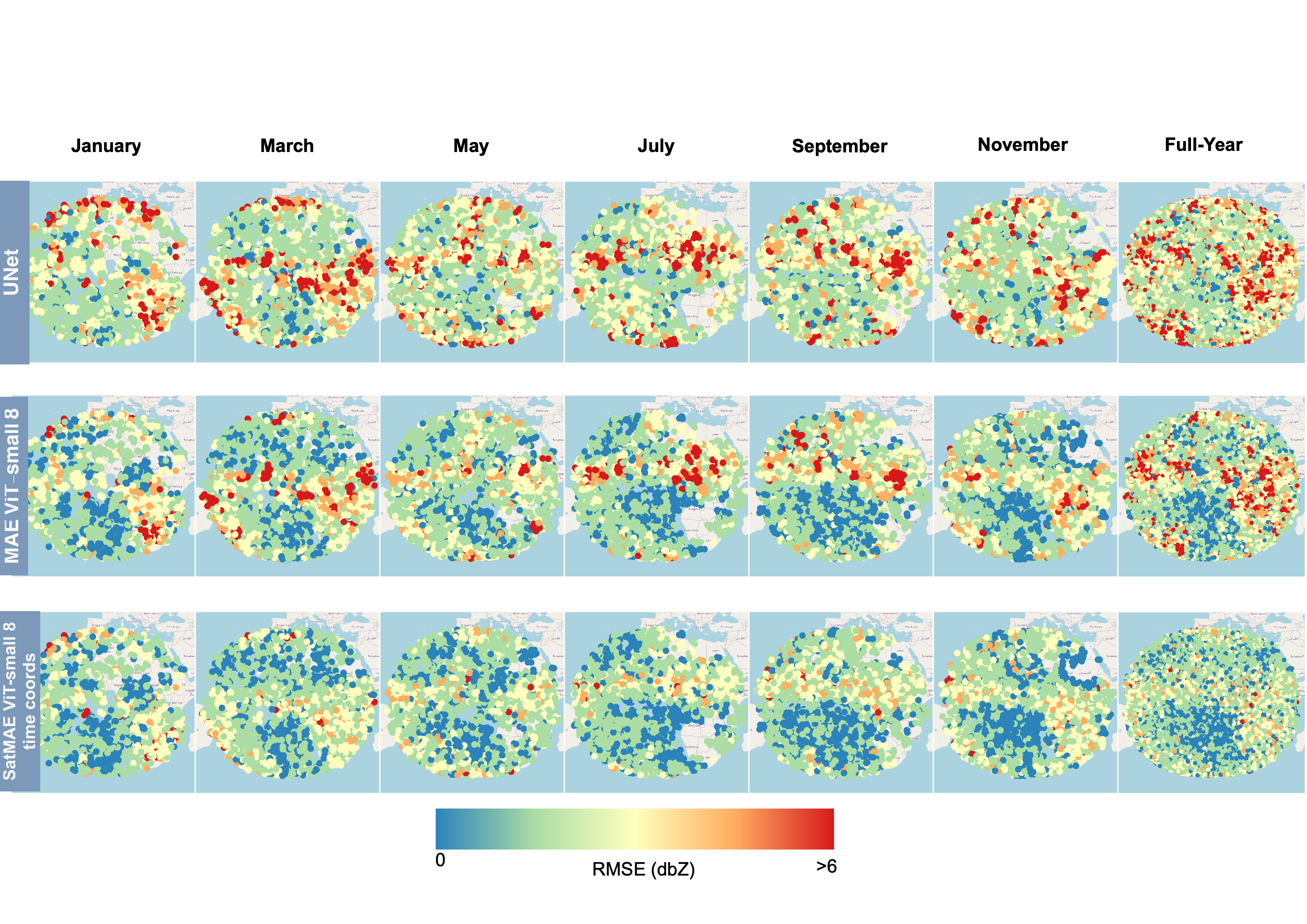}
    \caption{Monthly and yearly root-mean-square errors (RMSE) across MSG's field-of-view (one value per image-profile pair), comparing our U-Net, MAE, and SatMAE models. SatMAE with time and coords consistently improves overall prediction performance, especially in the tropical convection belt. Note that we show the prediction errors across our entire dataset (i.e. including training, validation, and test examples).}
    \label{Figure-RMSE-withencoding}
\end{figure}

\textbf{Geospatial Awareness.} Having demonstrated the benefit of self-supervised pre-training with ViT backbones that encode small-scale cloud features, we now compare the standard MAE to our space- and time-aware SatMAE model.
Looking at the perceptual quality of the predicted cloud profiles (Figure \ref{Figure-Predictions}), the MAE and SatMAE models lead to visually sharper reconstructions than the U-Net. For information, an example rendering of a predicted 3D cloud volume is shown in Figure \ref{Figure-3D-Reconstruction} in the Appendix, which shows good qualitative agreement with where clouds appear to be in the input MSG/SEVIRI image.
Table \ref{Table-Metrics-Comparison} shows that SatMAE achieves lower root-mean-square errors (RMSE) across the test set compared to both the U-Net and the MAE. 
SSIM and PSNR are similar across models, with a small improvement for SatMAE.
The values presented in Table \ref{Table-Metrics-Comparison} show the mean and standard deviation of the metrics across the test set, and do not represent actual model uncertainties. 
To gain more in-depth understanding on the performance of the models, we separated the contributions to the RMSE by cloud type (Table \ref{Table-RMSE-Cloudtypes} and Figure \ref{Figure-RMSE-Cloudtypes} in the Appendix). 
The largest errors are observed for Nimbostratus clouds and deep convection, i.e. rain and storm clouds with high radar reflectivities. 
For all cloud types except Nimbostratus, SatMAE achieves lower errors compared to both the baseline and the standard MAE.
Figure \ref{Figure-RMSE-withencoding} clearly demonstrates that the time and coordinate encodings positively impact performance, especially in the tropical convection belt, where cloud formations are larger and denser. Furthermore, Figure \ref{Figure-RMSE-satmae} in the Appendix shows that the coordinate encoding provides the most significant contribution to the model performance.
The improvement of the MAE model over the U-net, and that of the SatMAE over the MAE model are both calculated to have p-value of <0.001 showing the statistical significance of the improvement, when we take region into account, is very strong (see appendix \ref{appendix} for details on hypothesis test).

While U-Nets are effective in capturing local spatial relationships through their convolutional structure, they struggle with capturing global context. This limitation becomes apparent in the case of large-scale cloud formations, where long-range dependencies play a critical role. Transformers address this challenge by leveraging self-attention mechanisms, which enable them to model relationships across the entire input space. Moreover the use of custom encoding from coordinates and datetime, enables the model to distinguish between different geographic regions and temporal patterns, providing richer contextual information. The combination of this custom encoding with the smaller token sizes (8x8) allows SatMAE to achieve a more comprehensive understanding of global cloud structures and also to capture finer-grained details and local variations.

\section{Conclusions}

In this work, we used SSL via MAEs to pre-train models using unlabelled MSG/SEVIRI images, and then fine-tuned the models for 3D cloud reconstruction, using aligned pairs of MSG/SEVIRI and CloudSat/CPR radar reflectivity profiles.
In addition to the standard MAE implementation, we adapted a SatMAE model to encode the date, time, and location of the input data.
Compared to the U-Net, the current state-of-the-art on the task, our models showed improved performance across RMSE, PSNR, and perceptual quality.
While the improvements are moderate on average, visualization of prediction highlights that the SatMAE model achieves the lowest errors in the tropical convection belt, demonstrating its superior generalization.
This suggests that this learning set up has potential for larger gains for harder learning tasks such as 3D-cloud-type segmentation, helping further our goal of characterizing 3D cloud properties and contributing to the narrowing of climate model uncertainties. While our current work only explores reconstruction of CloudSat tracks for one year, future work will apply this method to longer time periods and ESA's EarthCARE data once available. This could enable the creation of long-term 3D cloud products, crucial for advancing our understanding of complex climate feedback mechanisms.



\section*{Acknowledgements}

This work has been enabled by FDL Europe | Earth Systems Lab (https://fdleurope.org) a public / private partnership between the European Space Agency (ESA), Trillium Technologies, the University of Oxford and leaders in commercial AI supported by Google Cloud, Scan AI and Nvidia Corporation.
\textcolor{blue}{
The material is based upon work under a programme of, and funded by, the European Space Agency. Any opinions, findings, and conclusions or recommendations expressed in this material are those of the author(s) and do not necessarily reflect the views of the European Space Agency.}

This work used rs-tools, a software package developed through the Trillium Technologies run Instrument-to-Instrument translation project (funded under NASA grant 22-MDRAIT22-0018/ 80NSSC23K1045).
E.P. also acknowledges the \textit{European Lighthouse of AI for Sustainability} (ELIAS) project under the Horizon  programme (Project ID. 101120237).
The authors would also like to thank the reviewers and experts that advised over the course of this project, including Paula Harder, Philip Stier, Duncan Watson-Parris, Timon Hummel, Kai Jeggle, Quentin Paletta, Mikolaj Czerkawski, Ronald Clarke, Chedy Raissi, and Yarin Gal.

\clearpage
\printbibliography

\clearpage


\appendix

\section{Appendix}
\label{appendix}

\paragraph{Further Training Details.}
We conducted our experiments on 1-2 NVIDIA V100 GPUs (16GB) via Google Cloud, with batch size 8 during pre-training and batch size 4 during fine-tuning. We used the Adam optimizer \cite{Kingma2015Adam:Optimization} with an initial learning rate of 0.00015, using backpropagation \cite{Rumelhart1986LearningErrors}.
Training was optimized via the Mean Squared Error (MSE) loss, while additional metrics like the Peak Signal-to-Noise Ratio (PSNR) and the Structural Similarity Index Measure (SSIM) \cite{1284395} were monitored. Checkpointing was used to save models with the lowest validation loss. Both pre-training and fine-tuning ran for up to 50 epochs. Regarding training times, the pretraining duration ranged from 2 to 6 days and was largely dependent on the patch size and the scale of the ViT backbone (ViT-small and ViT-base, see Table \ref{Table-Pretraining-Details}). Smaller patch sizes led to longer training times due to the increased number of tokens, while larger ViT architectures extended the duration due to the higher number of trainable parameters. For fine-tuning, training times ranged from 8 to 24 hours for the U-Net and MAE/SatMAE models with patch sizes of 16 and 8. With even smaller ViT patch size of 4, the training duration increased significantly, reaching up to 5 days.

\begin{table}[hb]
  \caption{
    Training details for pre-training of our Masked Autoencoder (MAE) and SatMAE models.}
  \label{Table-Pretraining-Details}
  \centering
  \begin{tabular}{lcccccccc}
  \toprule
    \multicolumn{1}{c}{Model} & \multicolumn{1}{c}{\begin{tabular}[c]{@{}c@{}}Time\\ encoding\end{tabular}} & \multicolumn{1}{c}{\begin{tabular}[c]{@{}c@{}}Spatial\\ encoding\end{tabular}} & \multicolumn{1}{c}{\begin{tabular}[c]{@{}c@{}}Token\\ Size\end{tabular}} & \multicolumn{1}{c}{\begin{tabular}[c]{@{}c@{}}ViT\\ size\end{tabular}} & \multicolumn{1}{c}{\# Tokens} & \multicolumn{1}{c}{\# Params} & \multicolumn{1}{c}{\begin{tabular}[c]{@{}c@{}}Training\\ time\end{tabular}} \\
   \midrule
    MAE & False & False & 8 & small & 1024 & 26.6 M & 22h 36 mins \\
    MAE & False & False & 8 & big & 1024 & 91.6 M & 1d 8h 38 mins \\
    MAE & False & False & 16 & small & 256 & 27.8 M & 17h 57 mins \\
    MAE & False & False & 16 & big & 256 & 93.3 M & 9h 22mins \\
    SatMAE & False & False & 8 & small & 1024 & 27.4 M & 22h 55 mins \\
    SatMAE & False & True & 8 & small & 1024 & 27.1 M & 22h 38 mins \\
    SatMAE & True & False & 8 & small & 1024 & 27.1 M & 22h 30 mins \\
    SatMAE & True & True & 8 & small & 1024 & 26.9 M & 23h 52 mins \\
    SatMAE & False & False & 16 & small & 256 & 28.2 M & 16h 57 mins \\
    SatMAE & False & True & 16 & small & 256 & 28.1 M & 17h 47 mins \\
    SatMAE & True & False & 16 & small & 256 & 28.1 M & 16h 38 mins \\
    SatMAE & True & True & 16 & small & 256 & 28.1 M & 15h 25 mins \\
    SatMAE & True & True & 16 & big & 256 & 93.8 M & 11h 30 mins \\
    SatMAE & True & False & 16 & big & 256 & 93.6 M & 11h 40 mins \\
    SatMAE & False & True & 16 & big & 256 & 93.6 M & 11h 19 mins \\
    SatMAE & False & False & 16 & big & 256 & 93.7 M & 11h 2 mins \\     
\end{tabular}
\end{table}

\begin{table}[hb]
  \caption{
    Training details for fine-tuning of our U-Net, MAE and SatMAE models.}
  \label{Table-Finetuning-Details}
  \centering
  \setlength{\tabcolsep}{4pt} 
  \begin{tabular}{lccccccccc}
  \toprule
    \multicolumn{1}{c}{Model} & \multicolumn{1}{c}{\begin{tabular}[c]{@{}c@{}}Time\\ encoding\end{tabular}} & \multicolumn{1}{c}{\begin{tabular}[c]{@{}c@{}}Spatial\\ encoding\end{tabular}} & \multicolumn{1}{c}{\begin{tabular}[c]{@{}c@{}}Token\\ Size\end{tabular}} & \multicolumn{1}{c}{\begin{tabular}[c]{@{}c@{}}ViT\\ size\end{tabular}} & \multicolumn{1}{c}{\# Tokens} & \multicolumn{1}{c}{\begin{tabular}[c]{@{}c@{}}Trainable\\ params\end{tabular}} & \multicolumn{1}{c}{Backbone} & \multicolumn{1}{c}{\begin{tabular}[c]{@{}c@{}}Training\\ time\end{tabular}} \\
   \midrule
    U-Net  & False & False & N/A & N/A & N/A & 1.9 M & N/A & 8h 50mins \\
    MAE & False & False & 8 & small & 1024 & 22.7M & from scratch & 22h 44mins \\
    MAE & False & False & 8 & small & 1024 & 345k & frozen & 16h 54mins \\
    MAE & False & False & 8 & small & 1024 & 22.7M & fine-tuned & 19h 45mins \\
    SATMAE & False & False & 8 & small & 1024 & 22.3M & fine-tuned & 19h 43mins \\
    SATMAE & True  & False & 8 & small & 1024 & 22.3M & fine-tuned & 19h 29mins \\
    SATMAE & False & True & 8 & small & 1024 & 22.3M & fine-tuned & 19h 8mins  \\
    SATMAE & True  & True & 8 & small & 1024 & 22.3M & fine-tuned & 19h 35mins \\
\end{tabular}
\end{table}

\clearpage

\textbf{Hypothesis test.} To show that the MAE-ViT-small 8 has superior performance to the U-net and the SatMAE-ViT- small 8 has superior performance to the MAE-ViT-small 8, for a random location we performed hypotheses tests with p-values which we report in section \ref{results}. The null hypothesis for this test is that while a given model may have a higher probability to obtain a higher RMSE for a given location, over the entire region under study, the expected probability  for any given location is 0.5, meaning on average, the models have equal performance. Under this null hypothesis the number of times model 1 is superior out of the
5,112 trials (the number observations in our test set) would be distributed as binomial random variable with parameters $n=46,752$ and $p=0.5$ assuming the outcome for each location was random, in which case we could construct a simple hypothesis test with a theoretical p-value. However,  it is reasonable to assume that $X_i\in\{1,2\}$, that describes which model is superior at location $i$ is correlated with $X_j\in \{1,2\}$ that describes which model is superior at location $j$ especially if $i$ and $j$ are close together. In order to avoid making specific assumption about the form of the spatial correlation we construct an empirical null distribution by taking the following steps:
\begin{enumerate}
\item construct $Y_i\in\{0,1\}$, $i=1,...,5,112$ that determines which model is superior at location $i$ in the following way $Y_i = \mathbbm{1}_{\{W_i<m\}}$ where:
\begin{itemize}
\item $\mathbbm{1}_x$ is the indicator function
\item $W_i=Z_i^1-Z_i^2$
\item $Z_i^k$ is the test rmse observed at location $i$ for model $k$ and,
\item $m$ is the test observed sample median of $W$ accross locations.
\end{itemize}
This ensures that the null distribution respects the null hypothesis assumption that for a random location the probability that any model is superior is 0.5, while also preserving the spatial correlation structure. 
\item We ideally would like to have many independent samples of $Y_i$ and then calculate, $P_i$, the proportion of times each model is superior at a location $i$. Since we only have one observation of each $Y_i$ we obtain a boostrap sample by using $Y_i$ and $Y_j$ for $j \in NN_{1000}(i)$ where $NN_{1000}(i)$ is the set of 1000 nearest neighbor locations to location $i$. Higher spatial correlation makes it more likely that, even though the probability that any given model is superior is 0.5, in an observed sample we observe a proportion far from the expected 0.5 (it increases the variance). The assumption is that the magnitude of the correlation at the smaller scales (1000 locations) is larger than the at larger scales (the scale of figure \ref{Figure-RMSE-withencoding}). If this holds our null distribution will be conservative meaning it has more variance than the actual null distribution, yielding valid p-values. 
\end{enumerate}
We then calculate the probability of obtaining the observed values for the proportion of times the MAE-ViT-small 8 has superior performance to the U-net ($0.85$) or a greater one, under the null distribution. Similarly, we calculate the probability of obtaining the observed values for the proportion of times the SatMAE-ViT- small 8 has superior performance to the MAE-ViT-small 8 ($0.73$) or greater, under the null. The probability of such extreme proportions of cases under the null distribution is less than 0.001 meaning both hypothesis tests are assigned a p-value of <0.001, and the null hypothesis can be rejected in both cases. 

\clearpage

\begin{figure}[ht!]
    \centering
    \includegraphics[width=\linewidth]{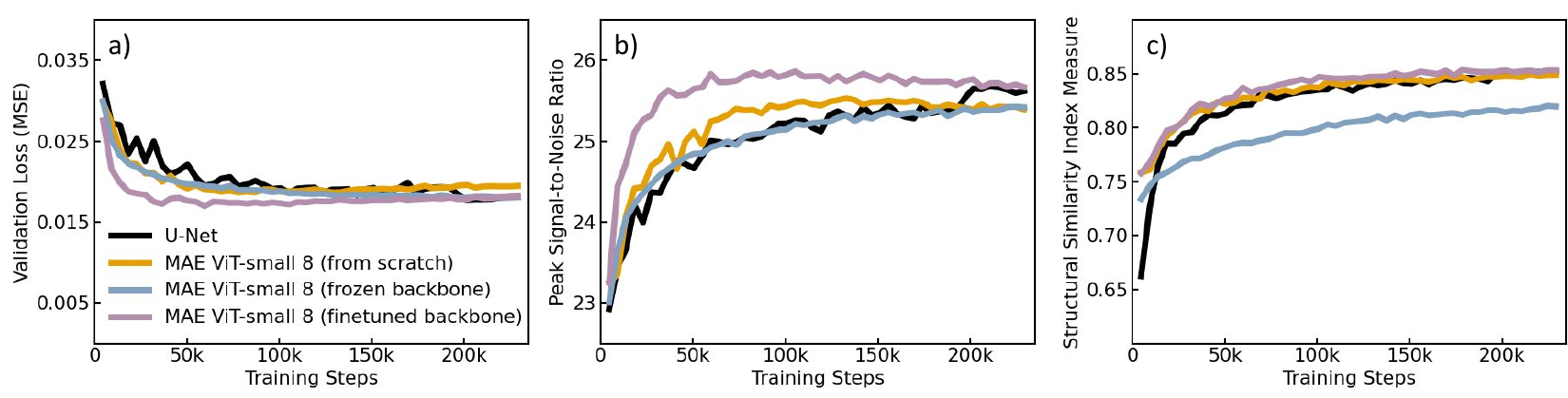}
    \caption{Mean squared error (MSE) loss, peak signal-to-noise ratio (PSNR), and structural similarity index measure (SSIM) as a function of training steps for our U-Net baseline and MAE models. The MAE was either trained from scratch, or fine-tuned using the pre-trained encoder. For the latter, we either froze or further fine-tuned the encoder weights.}
    \label{Figure-MAE-Comparison}
\end{figure}

\begin{figure}[htb]
    \centering
    \includegraphics[trim={0 5.5cm 0 7cm}, clip, width=0.95\linewidth]{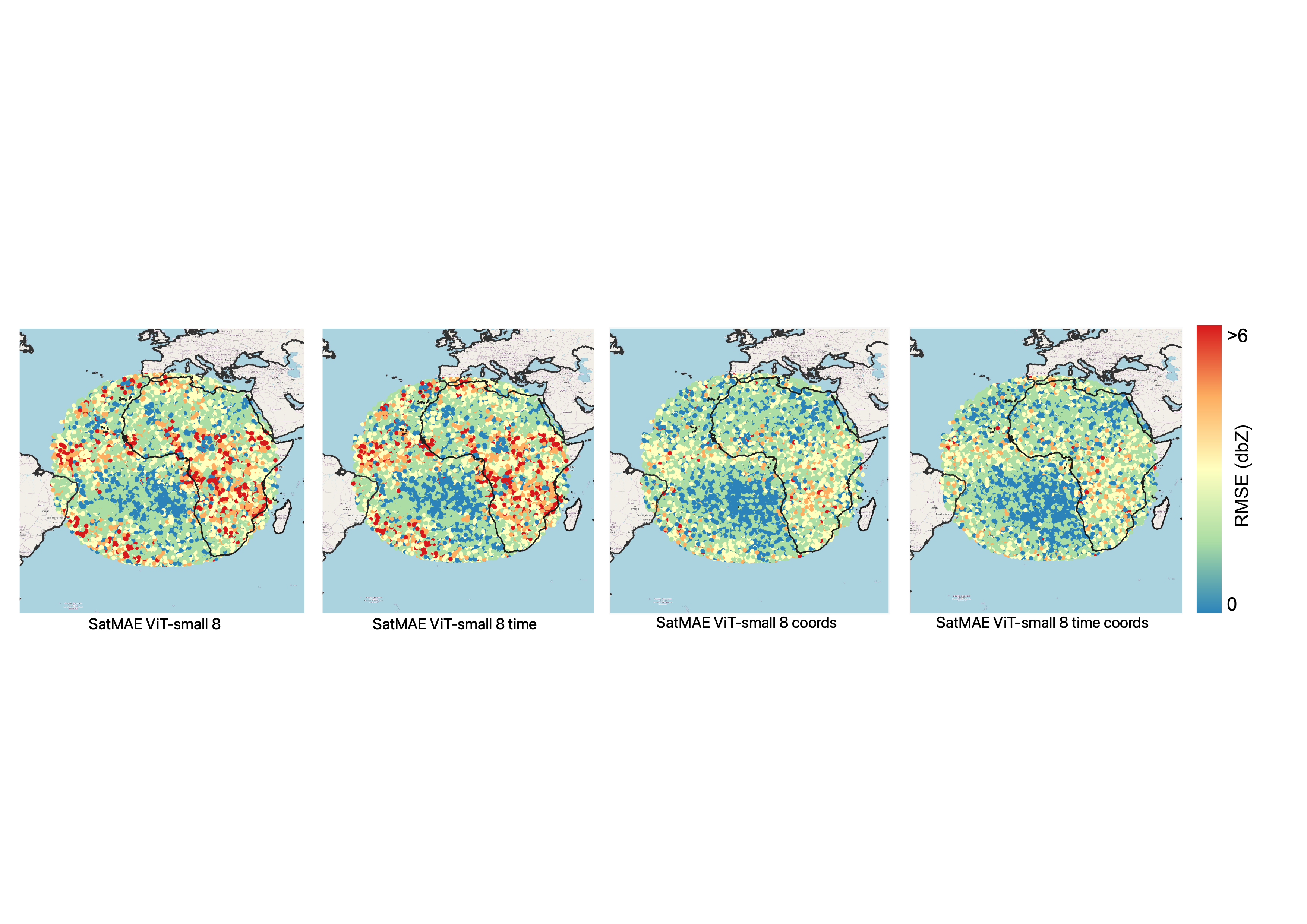}
    \caption{Root-mean-square errors (RMSE) across MSG's field-of-view (one value per image-profile pair), comparing different implementations of our SatMAE model. Adding coordinate encoding provides a strong improvement in prediction performance. 
    SatMAE with time and coords achieves the best performance overall, particularly in challenging areas such as the tropical convection belt. Note that we show the prediction errors across our entire dataset (i.e. including training, validation, and test examples).}
    \label{Figure-RMSE-satmae}
\end{figure}

\begin{figure}[ht!]
    \centering
    \includegraphics[width=\linewidth]{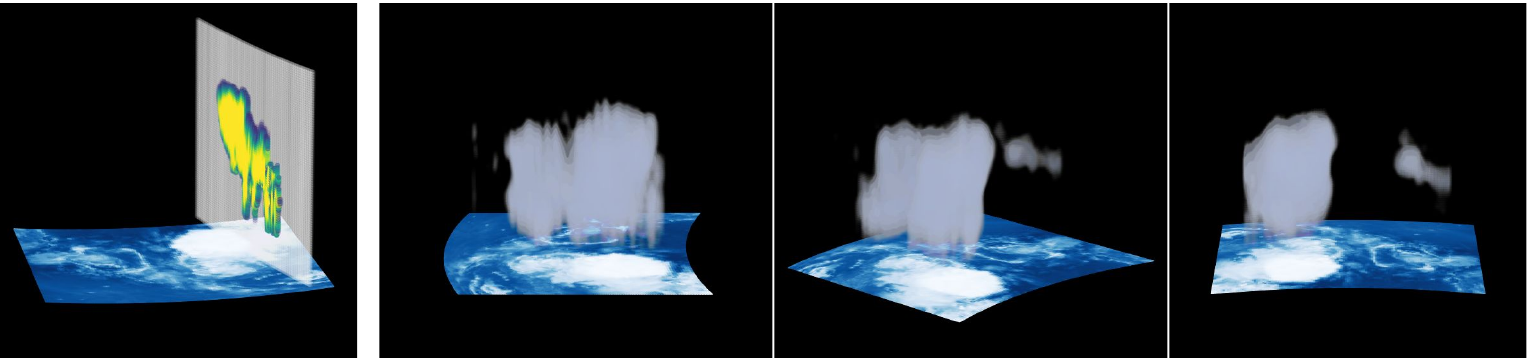}
    \caption{Example 3D cloud reconstPetersonruction. The left figure shows an MSG/SEVIRI input image with the aligned CloudSat/CPR track, while the three right figures show the predicted 3D volume rendered from different perspectives.}
    \label{Figure-3D-Reconstruction}
\end{figure}

\clearpage

\begin{table}[ht]
  \caption{Root-mean-square error (RMSE; in dBZ) as a function of cloud type for the different models. The values show the mean and standard deviation of all metrics across the test set.}
  \label{Table-RMSE-Cloudtypes}
  \centering
  \begin{tabular}{lccc}
    \toprule
    & U-Net & MAE ViT-small 8  & SatMAE ViT-small 8 (time + coords) \\
    \midrule
    No Cloud         & 2.59 $\pm$ 1.45  & 2.15 $\pm$ 1.43   & 2.13 $\pm$ 1.38                     \\ 
    Cirrus           & 5.22 $\pm$ 2.64  & 4.92 $\pm$ 2.44   & 4.77 $\pm$ 2.36                     \\ 
    Altostratus      & 10.69 $\pm$ 3.15 & 9.90 $\pm$ 2.73    & 9.77 $\pm$ 2.82                     \\
    Altocumulus      & 8.83 $\pm$ 4.63  & 8.44 $\pm$ 4.32   & 8.40 $\pm$ 4.28                      \\ 
    Stratus          & 5.32 $\pm$ 3.95  & 4.93 $\pm$ 3.77   & 4.83 $\pm$ 3.26                     \\ 
    Stratocumulus    & 8.76 $\pm$ 5.01  & 7.85 $\pm$ 4.49   & 7.52 $\pm$ 4.23                     \\ 
    Cumulus          & 7.35 $\pm$ 6.04  & 7.05 $\pm$ 5.88   & 6.99 $\pm$ 5.73                     \\ 
    Nimbostratus     & 14.19 $\pm$ 5.80  & 12.31 $\pm$ 4.86  & 12.53 $\pm$ 5.17                    \\
    Deep Convection & 11.23 $\pm$ 4.50  & 10.65 $\pm$ 3.82  & 10.34 $\pm$ 3.77                 \\
    \bottomrule
  \end{tabular}
\end{table}

\begin{figure}[htb]
    \centering
    \includegraphics[width=\linewidth]{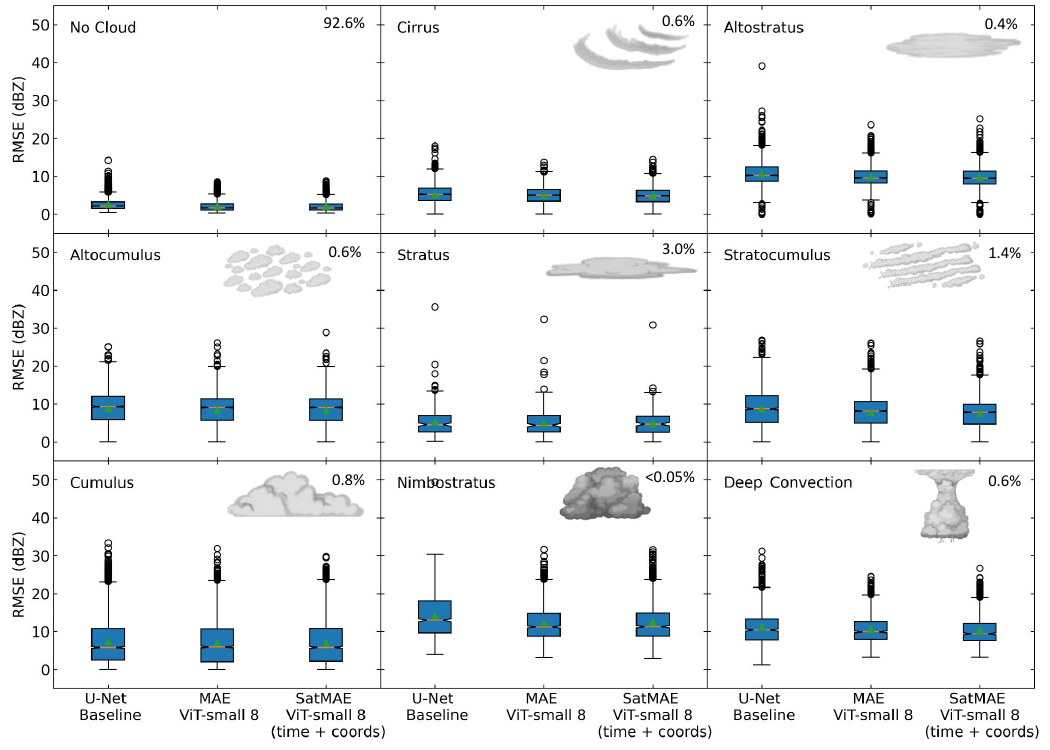}
    \caption{Root-mean-square error (RMSE; in dBZ) as a function of cloud type for the different models. The box and whisker plots summarize all samples in the test set. Little icons highlight what the different cloud types look like, and the percentages show the representation of each cloud type in our dataset. Even when limiting our dataset to CloudSat tracks that contain at least 20\% clouds, clear sky measurements are still vastly over-represented. Stronger radar reflectivities (i.e. from cumulus, nimbostratus, or deep convective clouds) lead to higher absolute errors. Furthermore, the sparse representation of nimbostratus clouds contributes to larger errors for this cloud type.}
    \label{Figure-RMSE-Cloudtypes}
\end{figure}

\end{document}